# Analysis of Online Toxicity Detection Using Machine Learning Approaches

Anjum[1], Rahul Katarya[2]

[1,2] Delhi Technological University, New Delhi, India-110042
anjum_2792@yahoo.com, rahulkatarya@gmail.com

**Abstract.** Social media and the internet have become an integral part of how people spread and consume information. Over a period of time, social media evolved dramatically, and almost half of the population is using social media to express their views and opinions. Online hate speech is one of the drawbacks of social media nowadays, which needs to be controlled. In this paper, we will understand how hate speech originated and what are the consequences of it; Trends of machine-learning algorithms to solve an online hate speech problem. This study contributes by providing a systematic approach to help researchers to identify a new research direction and elucidating the shortcomings of the studies and model, as well as providing future directions to advance the field.

## 1 Introduction

The internet is considered to be the best thing made by the humans of this era. It opened to the possibilities that no one ever imagined before. We could communicate with each other and share information effortlessly. The platform which has grown the most with it is social media. It is a platform where people can communicate with each other privately or publicly on the internet. The most used platforms in the world are Facebook 2.449 billion users, YouTube 2 billion users, WhatsApp 1.6 billion users, Twitter 340 million users[1]. These users belong to all kinds of age groups, sexuality, religion, ethnicity, culture, heritage, etc. People are entitled to free speech on these platforms, but this right is sometimes not used appropriately.

Along with them, the hate on these platforms has risen substantially. It could have various forms like cyberbullying, harassment, abusive language, discrimination, etc. which could take a toll on the mental health of the people facing it[2]. Hence, many researchers have studied the patterns of the spread of hate speech and have proposed ways to raise flags and as a result to control it. We have extracted some examples of online hate speech from Twitter, i.e., a person twitted while going to Africa, "Hope I don't get AIDS. Just kidding. I'm white!" This was a racist comment made in order to show that AIDS is a disease of race and considered them to be of the higher race, so they would not be affected by this disease[3].

**What is OHS and origin of It?**

Online hate speech (OHS) originated with the inception of the internet, but online social media (OSM) plays an essential role in the origin of online hate speech. Hate speech occurs when people or a group of people attack or use pejorative or discriminatory language against a community or an individual based on traits that define different aspects of one's individuality like origin, sexuality, ethnicity, religious background, socioeconomic strata, race, gender and various other attributes. When such an activity is carried through social media portals, blogs, creative content and other media available on the internet, it is classified as Online Hate Speech (OHS)[4].

Across the last decade, the advancements in technology, especially the internet as a medium of communication, have defined the functioning of the human race. The ease and reach provided by platforms like Facebook, Twitter and other social media giants have made it easier for people to communicate.

While all of these seem to be perks, it also comes with a lot of unaccounted influence that users have irrespective of their credibility of the matter. Moreover, the fact that users can anonymously tweet, snap or post content without actual validation and verification makes it very easy for people to hide their identity and yet be able to post whatever they want. Also, being able to express their opinions in a harsh or abusive manner and not have to justify them or formally argue validly makes it convenient to get away with hate speech[5].

All the above factors have it very easy for people to spread hate messages, abuse, pass sexist or racist comments and yet not be held accountable, irrespective of if they can or cannot be identified. Despite the existence of full-fledged cyber forces, even in the most advance and vigilant countries face huge difficulties in identifying hate speech and the spread of propaganda due to the vast networks that stretch across the internet. In a country like India, with limited resources, huge population and diversity in language as well as geography, it becomes even harder to locate and take appropriate action against offenders[6]. Fig. 1 shows examples of religious hate speech that exist in society and which was taken from Twitter [7].

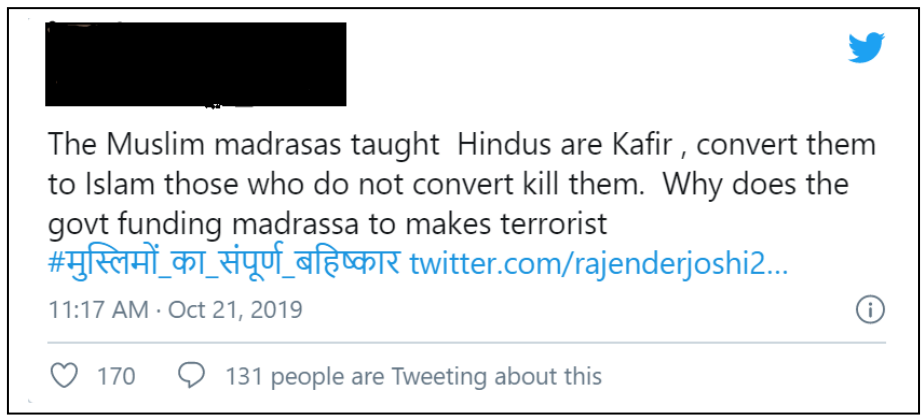

**Figure 1: Online religious hate speech**

**Perpetrator mission and consequences**

Since the internet, especially social media, is house to users of all ages, backgrounds and communities, it becomes easy for a perpetrator to be able to pit communities against each other to attain political supremacy, segment people of the basis of their differences by generating hate. They use such issues to attain political, social or economic powers or superiority like generating a vote bank for parties, influencing decisions in favor of the majority[8]. Many times, the offender uses it to condescend or belittle marginalized or minority communities and also not have to face them in real life. The offender can gain support from people with similar opinions and can stir propaganda or non-propaganda-based fights to scare or threaten people or communities and showoff their assumed "superiority"[9]. Another mission is to be able to influence young minds by instilling biases and extreme opinions. Moreover, they are exposed to an almost constant stream of information which they may not have the critical skills to filter and navigate. Hence, even without revealing their identity, a perpetrator can influence children through direct and indirect means[10].

There are three ways to look at the consequences of online hate speech. One is what the perpetrator has to suffer following online hate speech laws according to the country. Another is what the victim faces and lastly, how it affects the general society as a whole[10].

A victim is exposed to mental abuse as well as hampering of mental health, depending on how much the individual is affected by the hate speech. They can end up feeling insecure and unsafe. It can also lead to mental illnesses and disorders like depression, anxiety, sleep disorders. In general, it can be psychologically and mentally harmful and exhausting[11]. Although we have focused on how it can be very easy for an offender to use the internet as means of hate speech and get away with it, the current scenario, even though not perfect, does have laws to protect people against hate speech. In India, even though the fundamental rights provide citizens with free speech, hate speech is held as an exception to that law[12]. Therefore, and speech that is vehement, caustic, and sometimes unpleasantly sharp is held in violation of this exception, and the offender has tried accordingly. Moreover, there are various other legislation and self-regulatory mechanisms under which hate speech is negated[12].

The content of the paper is organized as follows; In section 1, we have researched about how hate speech originated and what are the consequences of it, in section 2 we discuss the previous state of the art models designed specifically for online hate speech, in section 3 we deeply explains the dataset used and the analysis of the machine learning classifier with respect to various features, section 4 presents all the experimented results and in section 5 we have discussed the conclusion of the paper.

## 2 Related Work

In this section, we also explored machine learning algorithms and techniques that are used to classify the sentences as hate and non-hate. As there are many techniques for the classification but to find the best features for the classification is an important task, sentiment analysis, dictionary-based approaches can be used to find the patterns in the sentences. The internet is on the boom, and along with it is social media. The number of daily active users in social media is more than ever. More and more people are connecting to it every day. Along with them, the hate on these platforms has risen substantially. It could have various forms like cyberbullying, harassment, abusive language, discrimination, etc. which could take a toll on the mental health of the people facing it. Hence, many researchers have studied the patterns of the spread of hate speech and have proposed ways to raise flags to control it. The studies have been mainly done using either machine learning or deep learning. Miró-Llinares et al. [13] made the use of the random forest technique to show how metadata patterns could help in creating an algorithm designed on a computer for detecting online hate speech. Agarwal et al. [14] used ML techniques like Decision Tree, Random Forest and Naïve Bayes along with lexicon features and NLP features to work on reducing the vagueness of intent of a post. They created a dataset using real-time data from the microbiologist website "Tumblr". Kiilu et al. [15] used Naïve Bayes along with N-grams and POS tagging to propose a method that could help in detecting and classifying hate speech on Twitter. They found out that the accuracy of the classification is directly proportional to the training data. Davidson et al. [16] used logistic regression along with Bag of Words and tf-idf to determine hate speech and abusive language and compare tweets from Standard American English and African American English. This paper proved the existence of racial bias in America. Lynn et al. [17] created a dataset in which they took definitions of the Urban Dictionary and classified them based on misogynistic and not misogynistic. Waseem et al. [18] used multi-task learning (MTL) along with Bag of Words, character N-grams and sub-word embeddings so that the model can learn subsidiary tasks along with the main task to give better results for the main task. Though it should be remembered that MLT does not guarantee an improvement inaccuracy. Almatarneh et al. [19] used various ML classifiers like Gaussian and Complement Naïve Bayes, Decision Tree, Random Forest, KNN and DL techniques like SVM and neural networks to compare various classifiers and find out which one gets the best accuracy in detecting hate speech on Twitter. Robinson et al. [20] used SVM and logistic regression along with unigrams, bigrams, trigrams and tf-idf to analyze feature selection and break the norm of feature engineering being significant. Omar et al. [21] used various ML and DL classifiers like Multinomial, Complement and Bernoulli NB, SVC, Nu SVC, Linear SVC, Logistic Regression, Decision Tree, SGD. Ridge, Perceptron, Nearest Centroid, RNN, CNN, to create a dataset that can be used for hate speech and abusive language detection in the Arabic language. Varade et al. [22] used LSTM along with word embeddings using gensim's word2vec method to detect hate speech in Hinglish language. Bisht et al. [23] used RNN, LSTM, Bi-LSTM neural networks and Stacked neural networks along with word embeddings to create a system that could help in separating hate speech from offensive language. Wani et al. [24] used various ML and DL classifiers like Decision Tree, Logistic Regression, Naïve Bayes, SVM, CNN and LSTM along with Char4-grams, Bigrams, Bag of Words, tf-idf to compare the accuracies of various classifiers for hate speech detection.

Concluding from the above works, it can be made clear that very few surveys and study papers have been published in the field of OHS. None of the papers had conducted various combinations of features on the balanced dataset. So, in this paper, we conducted a study of different features with the most used machine learning classifier. We also collected one combined dataset from the two unbalanced datasets, which gives better results than the previous state of the art.

## 3  Data and Evaluation

In this section, we have discussed the detailed description of the datasets, preprocessing steps, features selection and models used for the classification and evaluation. The dataset we have used consists of 41,684 tweets, which was almost equally divided into hate and non-hate categories. The basic architecture of the online hate speech detection model is shown in figure 2. First, the data was collected from the different online platforms, and that raw data needs to be preprocessed before analysis of it. After cleaning the data, all the words and sentences need to be converted into the vector so that we could perform some operations on it. To state the classification problem, it has essential to extract some relevant features from the dataset for better accuracy of the system. Then we could divide the dataset into training and testing split (80:20) ratio. And then we pass the data from the classifier, which calculates the results. The further description is stated in 3.1, 3.2, 3.3, and 3.4 sections.

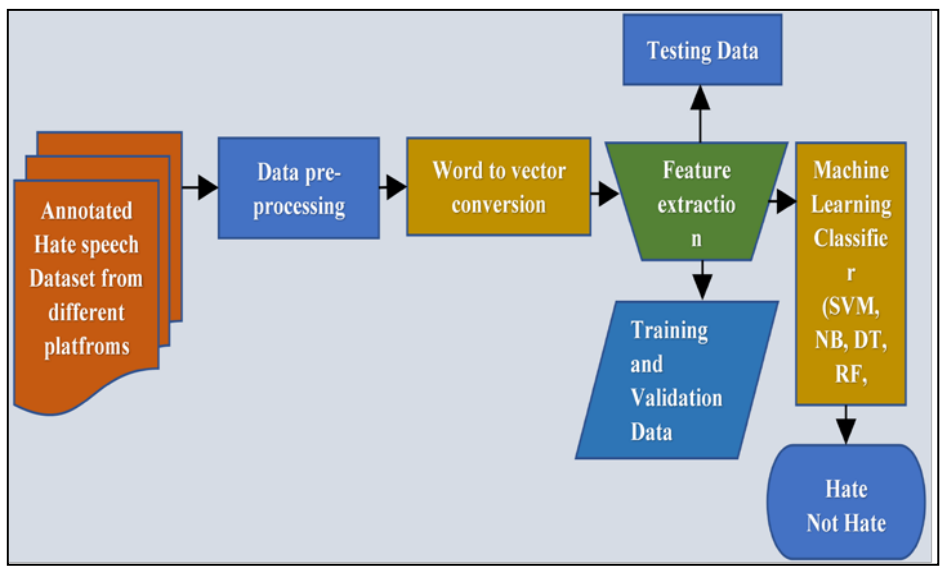

**Figure 2:** Machine learning model for hate speech detection

3.1 **Dataset:** Instead of using oversampling techniques to balance the given dataset, we have used two datasets, which give us a balanced dataset of around 41k English tweets.

   a) T Davidson: The dataset was collected from the GitHub website. This dataset comprises of 23,500 Tweets, which are further classified as hate, non-hate and offensive. We converted this dataset into a binary format in which offensive and non-hate were considered into one category ( non-hate); by this, we had two columns of hate and non-hate.

   b) Personal Attack (PA): This binary dataset was taken from the zenodo website, which was created under the European project for countering hate speech. For this dataset, we have collected tweets using Twitter API from Twitter.

   c) Combined dataset: We have explored various datasets for hate speech detection, but none of them found to be balanced. So, instead of using any oversampling techniques due to some disadvantages, we combined the above Two datasets shown in table 2.

   **Table 1:** Dataset Distribution

   | Class | T Devidson | PA | Combined |
   |---|---|---|---|
   | Hate | 1500 | 16,831 | **18,331** |
   | Non-hate | 4,000 | | |
   | offensive | 18,000 | 1,351 | **23,353** |
   | Total | 23,500 | 18,182 | **41,684** |

3.2 **Preprocessing:** As data always contains some sort of irrelevant details and noise, which does not play any significant role in the detection[25]. To preprocess the data, we used a pre-defined python nltk module which removed all the hashtags, usernames, emojis, URLs and numbers. We processed the data in the following ways steps:
   i. Removed HTML tags if any.
   ii. Removed square brackets or parenthesis if present.
   iii. Removed links from the tweets.
   iv. Removed '@' from the tweets.
   v. Removed stopwords.
   vi. Scrapped the extra white spaces from each sentence.
   vii. Finally, converting all sentences and words to lower case.

3.3 **Feature Selection:** To extract the most relevant features from the data is the most complex task of the classification. As we know, computer works on only numbers similarly, machine learning also works on the digits (binary format)[26]. So we converted the textual dataset set into vector format using tf-idf or sentence embeddings. We extracted three sets of features from the combined datasets. The first set of features obtained using tf-idf, the Second set of features obtained using

POS tagging and the third set were extracted using sentence embeddings. Bag of words (BOW) considers all the frequent words in the dictionary, which means it trains the classifier based on the frequency of the word (consider that word as a feature). But it does not consider the semantic of the sentence[26], that is why we performed our experiment by considering tf-idf vectorizer instead of BOW.

- **TF-IDF:** We used the inbuilt Sklearn library to implement tf-idf. It considers the importance of the word in the corpus and increases as the number of times that word occur in the document.
- **POS tag:** In order to evaluate the sematic or context of the sentence, we performed Part-of-the speech (POS) tag on each sentence.
- **Sentence embeddings:** We also performed the task using pre-trained Bidirectional Encoder Representations from Transformers (BERT) on the combined dataset. This model provides the ability to fine-tune the model for various downstream tasks[27]. To generate the embeddings, we used the TensorFlow library to load the tf module. BERT embeddings have a [CLS] token, which is a sentence level representation. The default representation is not very accurate, and it is better to fine-tune the classification model. So we use a library classed Sentence Transformers [8] to fine-tune the model. This library used pre-trained BERT models to generate sentence-level encodings. We have used the 'bert-base-nli-mean-tokens', which was trained on SNLI and MutilNLI dataset to create universal sentence embeddings. The embeddings are created by mean-token-pooling and generate a vector of size 768 for each sentence.

3.4 **Models**: We have experimented on four most used machine learning classifiers, i.e., logistic regression, support vector machine (SVM), Multinomial Naïve Bayes and random forest (RF). Each set of features has experimented with each classifier. Logistic regression was a simple classification algorithm that has the advantage of showing well-calibrated probabilities that other algorithms fail to do. Another algorithm that we used is the Random Forest. It is an ensemble-based learning algorithm that combines the results of various decision trees and takes out their best features to give better results. It has the advantage of having low bias and moderate variance, along with having significantly less training time. Another algorithm is Naïve Bayes. The Naïve Bayes that we have used for our work is Multinomial Naïve Bayes. It has the advantage of high scalability and insensitivity towards irrelevant features. We have also used SVM for our work. The particular technique that we have used here is Linear SVC. It is also a classification algorithm that has the advantage of being more memory efficient than other algorithms and also it has a better ability to solve higher-dimensional problems.

# 4 Results

We performed several experiments to analyze and evaluate the performance of each four classifiers and obtained good results. In the first experiment, we have covered the top 400 words that have the highest tf-idf score values. The accuracy, precision, recall and f-score were calculated for each classifier. We found that as compared to all four classifiers, linear SVC(SVM) gives the best results, but multinominal outperformed all the classifiers in terms of precision.

### 4.1 Experiment 1: Using TFIDF only

In the first experiment, we used only tf-idf as a feature. We applied it on the combined dataset column that contained the tweets. Using this, we created the feature vectors of the tweets. Since it is not feasible to take all the features, we took a maximum of 500 features tp experimented. After extracted the features, we transformed the X_train and X_test according to these vectors as stored them in tfidf_train and tfidf_test, respectively. Then, we applied the models on tfidf_train. We used all the models mentioned above and applied the evaluation matrices to checked out the results of this experiment. The results are tabulated in Table 3.

**Table 2:** Classification using Tf-Idf feature set

| Classifier | Accuracy | Precision | Recall | f1-score |
|---|---|---|---|---|
| **Logistic Regression** | 0.922874 | 0.909140 | 0.915847 | 0.912481 |
| **Random Forest** | 0.915797 | 0.902778 | 0.905738 | 0.904255 |
| **SVM (Linear SVC)** | **0.922994** | 0.907618 | **0.918033** | **0.912795** |
| **Multinomial Naïve Bayes** | 0.897445 | **0.923332** | 0.835792 | 0.877384 |

### 4.2 Experiment 2: Using TFIDF on POS tags

For the second experiment, we first created POS tags of the preprocessed data and stored in a new column with the name "tagged". After that, we used tf-idf on this column. The transformed values after vectorization were stored in tf-idf. Then we combined the features of tf-idf on tweets with features of tf-idf on tagged words using POS tagger and stored them in the tfidf_comb variable. Then the dataset split into training and testing data to run our models on them. We used all the models as mentioned above and evaluate the results of this experiment. We presented the results in Table 4.

**Table 3:** Classification Using TFIDF on POS tags

| Classifier | Accuracy | Precision | Recall | f1-score |
|---|---|---|---|---|
| **Logistic Regression** | **0.923233** | 0.910996 | 0.914480 | **0.912735** |
| **Random Forest** | 0.910159 | 0.900412 | 0.894262 | 0.897327 |
| **SVM (Linear SVC)** | 0.922753 | 0.908672 | **0.916120** | 0.912381 |
| **Multinomial Naïve Bayes** | 0.899724 | **0.921996** | 0.842896 | 0.880674 |

### 4.3 Experiment 3: Using Sentence Embeddings

For the third experiment, we took the help of Universal Sentence Encoder. First of all, we used it to create embeddings of the tweets. Then, we combined these embeddings with the combined tf-idf features of tweets and POS tags. After this, we split these embeddings into training and testing data to run our models on them. We used all the aforementioned models except Naïve Bayes and applied the evaluation matrices to check out the results of this experiment. We did not use Naïve Bayes as it fails to work with the negative values that were created because of the Universal Sentence Encoder. The results are presented in Table 5.

**Table 4:** classification Using Sentence Embeddings

| Classifier | Accuracy | Precision | Recall | f1-score |
|---|---|---|---|---|
| Logistic Regression | 0.917836 | 0.914923 | 0.896175 | 0.905452 |
| Random Forest | 0.910759 | **0.920415** | 0.872131 | 0.895623 |
| SVM (Linear SVC) | **0.918796** | 0.913043 | **0.900819** | **0.906890** |

## 5. Conclusion

Like the online social media increase, the acts of online hate speech can lead to the normalization of discrimination, prejudice and promote ideas like supremacy, majoritarianism and can instill or reinstil unjust societal norms like the caste system. It can also lead to general disharmony and generate hate and toxicity across different channels of communication and social media. Moreover, if the situation escalates, it can end up in real-life violence and riots, which only make things worse. Many social media like Twitter, Facebook, etc. are looking for an artificial intelligence solution to stop such kinds of activities on OSM. In order to contribute, we had performed some experiments, where we found that SVM gives better accuracy in two out of three cases, but it gives high recall rates in all the cases means most of the time SVM gives relevant results compared to other classifiers. We also found that Tf-Idf features give the better feature set as compared to the other competitor. In the future, we can create a balanced dataset of Hindi and English language.